\documentclass{article}
\usepackage{amsmath} 
\usepackage{amssymb} 
\usepackage{graphicx} 
\usepackage{booktabs}
\usepackage{multirow}
\usepackage[most]{tcolorbox}
\usepackage{enumitem}
\usepackage{url}
\usepackage{hyperref}
\usepackage{authblk}
\usepackage{float} 

\title{Benchmark Success, Clinical Failure: \\
When Reinforcement Learning Optimizes for Benchmarks, Not Patients}

\author[1,2,3]{Armin Berger\thanks{These authors contributed equally and share first authorship.}}
\author[1,2,3]{Manuela Bergau$^*$}
\author{Helen Schneider}
\author[1]{Saad Ahmad}
\author[4,5]{Tom Anglim Lagones}
\author[6]{Gianluca Brugnara}
\author[6]{Martha Foltyn-Dumitru}
\author[6]{Kai Schlamp}
\author[6]{Philipp Vollmuth}
\author[1,2]{Rafet Sifa}

\affil[1]{Fraunhofer IAIS, Germany}
\affil[2]{University of Bonn , Germany}
\affil[3]{Lamarr Institute, Germany}
\affil[4]{Department of Health Queensland, Australia}
\affil[5]{Griffith University, Australia}
\affil[6]{University Hospital Bonn, Germany}

\date{December 2025}

\begin{document}

\maketitle

\begin{abstract}
Recent Reinforcement Learning (RL) advances for Large Language Models (LLMs) have improved reasoning tasks, yet their resource-constrained application to medical imaging remains underexplored. We introduce ChexReason, a vision-language model trained via R1-style methodology (SFT followed by GRPO) using only 2,000 SFT samples, 1,000 RL samples, and a single A100 GPU. Evaluations on CheXpert and NIH benchmarks reveal a fundamental tension: GRPO recovers in-distribution performance (23\% improvement on CheXpert, macro-F1 = 0.346) but degrades cross-dataset transferability (19\% drop on NIH). This mirrors high-resource models like NV-Reason-CXR-3B, suggesting the issue stems from the RL paradigm rather than scale. We identify a generalization paradox where the SFT checkpoint uniquely improves on NIH before optimization, indicating teacher-guided reasoning captures more institution-agnostic features. Furthermore, cross-model comparisons show structured reasoning scaffolds benefit general-purpose VLMs but offer minimal gain for medically pre-trained models. Consequently, curated supervised fine-tuning may outperform aggressive RL for clinical deployment requiring robustness across diverse populations.
\end{abstract}

\section{Introduction}

Recent work demonstrates that reinforcement learning (RL) can substantially improve large language model performance, particularly in settings with a clear reward signal and automatically verifiable outcomes (e.g., mathematics and code generation; see, e.g., DeepSeek-R1). However, it remains less clear how reliably these gains transfer to problems with weaker or more subjective supervision, such as free-form natural language generation and multimodal inputs. In this work, we investigate whether R1-style training, which combines supervised fine-tuning (SFT) with Group Relative Policy Optimization (GRPO), can enhance multilabel chest X-ray classification in small vision-language models under severe resource constraints. We focus on chest X-ray diagnosis because it represents a clinically critical task where radiologists value both hard diagnostic labels for rapid assessment and accompanying reasoning traces to establish trust in model outputs. Moreover, chest X-rays benefit from large publicly available datasets with multilabel annotations that provide natural reward signals for reinforcement learning.
\bigskip
While recent work has explored R1-style reasoning for medical visual question answering, multilabel chest X-ray classification remains less studied. A notable exception is NVIDIA's NV-Reason-CXR-3B, which utilizes extensive synthetic data and compute. Our work contrasts with this high-resource approach by examining R1-style training under extreme constraints: 50 times less training data and 4 times less compute. This setting is particularly relevant for practitioners who lack large-scale annotation pipelines or extensive infrastructure but still seek to leverage reasoning-guided training for improved diagnostic performance. Our work makes three primary contributions.

\begin{itemize}
    \item \textit{Low-Resource R1-Style Training:} We present ChexReason, trained with only 2,000 SFT and 1,000 RL samples on a single A100 GPU, demonstrating that R1-style training is feasible without extensive resources.
    
    \item \textit{Instruction Format Sensitivity:} Cross-model analysis reveals that optimal instruction format depends on medical pre-training: structured medically informed reasoning scaffolds benefit general-purpose VLMs while providing minimal gain for domain-specialized models.
    
    \item \textit{Benchmark-Transferability Trade-off:} GRPO improves CheXpert performance (+23\%) but degrades NIH transferability ($-$19\%), mirroring NV-Reason-CXR-3B failures and suggesting a paradigm-level issue.
    
    \item \textit{Generalization Paradox:} The SFT checkpoint uniquely improves on out-of-distribution data, indicating teacher-guided traces capture more generalizable features than reward-optimized outputs.
\end{itemize}

\section{Related Work}

Recent advancements in large language models have spurred significant interest in applying reinforcement learning (RL) and chain-of-thought (CoT) reasoning to medical vision-language models (VLMs), a trend motivated by the success of general-domain approaches like DeepSeek-R1. Consequently, several studies have explored R1-style reasoning recipes for medical visual question answering (VQA). For instance, MedVLM-R1 \cite{medvlm_r1_2025} utilizes GRPO to improve VQA accuracy significantly across MRI, CT, and X-ray benchmarks. Similarly, the introduction of Med-R1 \cite{med_r1_2025} demonstrated that RL can enhance generalization and reliability across eight imaging modalities, notably outperforming much larger models. To address computational constraints, RARL \cite{rarl_2025} employs Low-Rank Adaptation (LoRA) to improve reasoning on VQA tasks, while other works such as GMAI-VL-R1 \cite{gmai_vl_r1_2025} have harnessed RL for multimodal reasoning on large-scale datasets. Furthermore, recent research has explored eliciting medical reasoning from base models using verifiable rewards without explicit supervision, yielding improvements in out-of-distribution generalization \cite{med_rlvr_2025}.

While VQA-focused approaches have proliferated, the application of R1-style training to multilabel chest X-ray classification remains less explored. A pioneering effort in this specific domain is NV-Reason-CXR-3B \cite{myronenko_nv-reason-cxr_2025}, which applies reasoning with GRPO to the CheXpert ontology. This model produces radiologist-style stepwise reasoning traces using a two-stage training pipeline on a Qwen2.5-VL-3B-Instruct backbone, scaling supervision via synthetic data derived from radiology reports. This paper by Nvidia, despite its similarity to our approach, was developed independently from the research of this paper and published weeks earlier. Complementing this methodology, ChestX-Reasoner \cite{fan_chestx-reasoner_2025} leverages process supervision mined from clinical reports and introduces RadRBench-CXR, a benchmark designed to evaluate reasoning quality in chest X-ray interpretation.

Within the broader chest X-ray domain, various approaches have been developed to tackle diagnosis and report generation. Interpretability has been a key focus; X-Ray-CoT \cite{ng_x-ray-cot_2025}, for example, proposes a framework for diagnosis using CoT reasoning that achieves high balanced accuracy on the CORDA dataset. In the realm of conversational models, RadVLM \cite{radvlm_2025} excels in multitask capabilities including classification, localization, and captioning. Other unified frameworks include MedRAX \cite{fallahpour_medrax_2025}, an agentic framework that integrates state-of-the-art analysis tools with multimodal LLMs.

Report generation has also benefited from targeted technical improvements designed to enhance clinical accuracy. To address the issue of hallucinations, particularly regarding prior exams, Direct Preference Optimization (DPO) has been successfully employed to reduce hallucinated lines significantly \cite{dpo_hallucination_2024}. Additionally, joint localization and classification have been explored using combinations of Llama-2-Chat and BiomedCLIP \cite{litegpt_2024}. These modeling advances are supported by evolving evaluation frameworks, such as CXPMRG-Bench \cite{wang_cxpmrg-bench_2024} for report generation and ReXrank \cite{zhang_rexrank_2024}, which establishes standardized metrics across multiple datasets.

Beyond reasoning-based approaches, classification and structured prediction have been addressed through diverse strategies. Multi-task networks like CLN \cite{okolo_cln_2025} have been developed for simultaneous localization and classification, while other studies have utilized small LLMs to automate the transformation of unstructured radiology reports into structured labels \cite{abdullah_automated_2025}. Collaboration between clinicians and VLMs has also been investigated to enhance report generation \cite{tanno_collaboration_2025}. Furthermore, competitive results on validation sets have been achieved through federated learning \cite{lotfinia_boosting_2025}, CLIP-based zero-shot learning with text embeddings \cite{bhardwaj_enhancing_2025}, and contrastive learning with partial label overlap loss \cite{jang_significantly_2024}. Finally, RL fine-tuning continues to be refined for medical VQA, with recent works focusing on visual grounding \cite{gemex_thinkvg_2025}, the consistent benefits of GRPO \cite{effective_rl_medvqa_2025}, and aligning models with clinical preferences \cite{mmedpo_2025}.

These medical-specific developments are deeply informed by broader advances in visual reasoning. Innovations such as vision-guided RL for human-free alignment \cite{vision_r1_2025} and the optimization of visual reasoning via CoT responses \cite{reason_rft_2025} provide the foundational techniques that are currently being adapted to enhance performance in medical imaging tasks.

\section{Methodology}

This section describes our methodology for training a vision-language model to perform multilabel classification of chest X-ray images with accompanying chain-of-thought reasoning traces. Our approach consists of two main stages: supervised fine-tuning (SFT) using high-quality reasoning traces, followed by reinforcement learning with Group Relative Policy Optimization (GRPO) to further refine performance. A central goal is to elicit deliberate reasoning behaviour, what might be termed ``level 2 thinking,'' rather than shallow pattern matching, through the reinforcement learning process.
For our experiments, we employed MedGemma-4B, a multimodal vision-language model that processes images normalized to $896 \times 896$ pixels into 256 image tokens. We selected this model for its extensive pretraining on biomedical data and its domain-adapted CLIP encoder for visual processing. To evaluate generalization across different architectures, we also conducted comparative experiments using Qwen2.5-VL-3B-Instruct. All training runs were executed on a single NVIDIA A100 80GB GPU.

\subsection{Dataset}

For training, we utilized the MIMIC-CXR-JPG dataset \cite{johnson_mimic-cxr-jpg_2019}, which contains 377,110 chest X-ray images associated with 227,835 imaging studies from 65,379 patients \cite{hu_medical-cxr-vqa_2025, hu_interpretable_2024, goldberger_physiobank_2000}. We selected only datapoints with antero-posterior or postero-anterior views and considered only positive CheXpert labels. The selection process employed a penalty-based greedy iterative sampler designed to score candidates by favoring images that reduce deficits for under-represented labels while heavily penalizing choices that would exacerbate the over-representation of abundant labels. This strategy ensures each label is present in at least 5\% of the samples. Using this approach, we generated 2,000 samples for the Supervised Fine-Tuning (SFT) step and 1,000 samples for the Reinforcement Learning (RL) step, with no data overlap between the two subsets.

For the reasoning traces dataset leveraged in the SFT stage, each of the 2,000 instruction-following examples was generated using Gemini 2.5 as a teacher model. The teacher model was provided with ground-truth labels during generation but was instructed to produce rationales as if reasoning independently. We developed this dataset in collaboration with doctors and radiologists from the University Clinic Bonn (UKB) and Queensland Health to mirror the medical reasoning process radiologists undergo when examining chest X-rays. Our development process involved sampling 100 datapoints from the 2,000 selected for SFT training and generating corresponding examples using medically informed reasoning. These datapoints were reviewed by radiologists at the UKB, and the prompting strategy was incrementally refined to deliver medically sound reasoning traces.

Furthermore, we extracted an additional 500 samples at random from the same dataset \cite{johnson_mimic-cxr-jpg_2019} to serve as a validation set. Using this validation set, we evaluated the performance of different prompting strategies and their respective effects during SFT training. We also used this set to cross-compare the response to training of two student models: MedGemma-4B and Qwen2.5-VL-3B-Instruct.

For evaluation, we employed two out-of-distribution test sets. The first is a subset of the CheXpert dataset \cite{Irvin2019CheXpert}, a large-scale collection from Stanford Hospital comprising 224,316 chest radiographs from 65,240 patients. These are labeled for 14 common observations using an automated rule-based labeler that explicitly captures uncertainty from radiology reports. The second is a subset of the NIH Chest X-ray dataset \cite{bahaaeldin2024nihchestxray14}, which contains 112,120 frontal-view X-rays from 30,805 unique patients at the NIH Clinical Center, annotated with 14 disease labels derived via natural language processing of associated reports. Unlike CheXpert, the NIH Chest X-ray 14 dataset does not explicitly label uncertainty. From these datasets, we utilized 518 observations from CheXpert and 488 observations from NIH for testing. We reduced the NIH label set to nine pathology categories to match a subset of the CheXpert labels: Atelectasis, Cardiomegaly, Consolidation, Edema, Lung Lesion, No Finding, Pleural Other, Pneumonia, and Pneumothorax.

\subsection{Training}

We divide training into two stages: supervised fine-tuning (SFT) followed by Group Relative Policy Optimization (GRPO). Throughout this section, we use the following notation: input X-ray images are denoted as $x$, generated text sequences (comprising reasoning and predictions) as $t$, ground truth labels as $Y$, and predicted labels as $\hat{Y}$ from the 14-label CheXpert lexicon. The model outputs text in a structured format \textbf{\texttt{\{analysis: \dots, conclusion: \dots\}}}, from which we extract predicted labels using a rule-based parser.

\subsubsection{Supervised Fine-Tuning}

In the SFT stage, we train the model to generate teacher-annotated reasoning traces from the 2,000-sample training dataset. The objective minimizes the negative log-likelihood of expert traces $t^*$ given input X-ray images $x$:
\begin{equation}
\mathcal{L}_{\text{SFT}}(\theta) = -\mathbb{E}_{(x, t^*) \sim \mathcal{D}_{\text{SFT}}} \left[ \sum_{j=1}^{|t^*|} \log \pi_\theta(t^*_j \mid x, t^*_{<j}) \right],
\end{equation}
where $t^*_j$ denotes the $j$-th token in the expert trace, and each trace was generated by Gemini 2.5 provided with ground-truth labels but instructed to produce rationales as if reasoning independently, resulting in clinician-validated reasoning patterns.

We performed SFT on two vision-language models: MedGemma-4B~\cite{medgemma2024} and Qwen2.5-VL-3B-Instruct ~\cite{qwen2025qwen2.5vl3b}, using the Accelerate library~\cite{accelerate}. To enable parameter-efficient training, we employed Low-Rank Adaptation (LoRA)~\cite{hu2022lora} with rank 16, scaling factor 32, and dropout rate 0.1, applied exclusively to the language model's attention and feed-forward projection layers, while keeping the vision encoder frozen throughout training. Both models were trained for up to 6 epochs with early stopping, patience of 2 epochs based on validation loss, using the AdamW optimizer with a learning rate of $5 \times 10^{-5}$, cosine annealing schedule with 5\% warmup, weight decay of 0.01, and gradient clipping at norm 5.0. We used an effective batch size of 8 achieved through gradient accumulation with a per-device batch size of 1, and reserved 10\% of the data for validation. Training was conducted in BFloat16 precision with gradient checkpointing enabled for memory efficiency. Critically, we applied assistant-only loss masking, where loss computation was restricted to the assistant's response tokens by masking the user prompt, image placeholder tokens, and padding tokens. For visual token processing, we controlled image resolution by constraining visual tokens to the range of 256--512 tokens per image via the processor's minimum and maximum pixels parameters.

\subsubsection{Group Relative Policy Optimization}

Following SFT, we fine-tune the model using Group Relative Policy Optimization (GRPO)~\cite{shao2024deepseekmath} on the 1,000-sample RL dataset. GRPO obviates the need for an additional value function approximation and instead uses the average reward of multiple sampled outputs as the baseline. Specifically, for each input X-ray $x$, GRPO samples a group of text completions $\{t_1, t_2, \cdots, t_G\}$ from the old policy $\pi_{\theta_{\text{old}}}$ and optimizes the policy model by maximizing the following objective:

\begin{align}
\mathcal{J}_{GRPO}(\theta) &= \mathbb{E}_{\substack{x \sim \mathcal{D}_{\text{RL}}, \, \{t_i\}_{i=1}^G \sim \pi_{\theta_{\text{old}}}(t \mid x)}} \Bigg[ \frac{1}{G} \sum_{i=1}^G \frac{1}{|t_i|} \sum_{j=1}^{|t_i|} \min \big[ r_{i,j}(\theta) \hat{A}_i, \nonumber \\
&\qquad\qquad \text{clip}(r_{i,j}(\theta), 1{-}\varepsilon, 1{+}\varepsilon) \hat{A}_i \big] - \beta \mathbb{D}_{KL}(\pi_\theta \| \pi_{ref}) \Bigg],
\end{align}

where $r_{i,j}(\theta) = \frac{\pi_\theta(t_{i,j} \mid x, t_{i,<j})}{\pi_{\theta_{\text{old}}}(t_{i,j} \mid x, t_{i,<j})}$ is the token-level importance ratio, $\hat{A}_i$ is the sequence-level advantage calculated based on relative rewards within each group, and $\varepsilon$ and $\beta$ are hyper-parameters controlling the clipping bounds and KL divergence penalty, respectively. This group-relative approach aligns well with the comparative nature of reward models, which are typically trained on comparison datasets. Furthermore, instead of adding a KL penalty to the reward, GRPO regularizes by directly adding the KL divergence term between the trained policy and the reference policy $\pi_{ref}$, the SFT checkpoint, to the objective ~\cite{shao2024deepseekmath}.

\textbf{Stabilizing Training Against Mode Collapse.} In early training runs, we observed training instability characterized by sharp drops in policy entropy, repetitive outputs, and rapid performance degradation on validation metrics, symptoms consistent with mode collapse in RL fine-tuning of LLMs~\cite{zheng2025stabilizing}. To address this, we incorporated several stabilization mechanisms informed by recent work on stable RL training for language models. Specifically, our formulation includes: (1) token-level importance sampling correction via the ratio $\pi_\theta / \pi_{\theta_{\text{old}}}$ to account for policy updates between rollout and training steps, (2) clipped policy updates to constrain policy staleness and prevent aggressive parameter changes, and (3) explicit KL divergence regularization against the reference policy to maintain proximity to the SFT checkpoint~\cite{zheng2025stabilizing}. These modifications eliminated the collapse behavior observed in preliminary experiments across all training runs.

For each training sample, we generated 4 completions at temperature 0.8 with nucleus sampling (top-$p=0.95$). The optimization balances reward maximization against a KL divergence penalty ($\beta=0.15$) to prevent excessive deviation from the SFT checkpoint. We employed the Dr.~GRPO loss normalization~\cite{liu2025understanding} to eliminate length bias, with asymmetric PPO clipping bounds $[0.15, 0.22]$ for stable policy updates.

\textbf{Reward Functions.} We developed two alternative reward functions with contrasting design philosophies and evaluated their performance in separate training runs. The \emph{hard reward} enforces strict format compliance combined with label prediction accuracy:
\begin{equation}
r_{\text{hard}}(t, Y) = \begin{cases}
J(Y, \hat{Y}) - \lambda_{\text{len}} & \text{if } t \text{ is valid and } |t| < 250 \\
J(Y, \hat{Y}) & \text{if } t \text{ is valid and } |t| \geq 250 \\
0 & \text{otherwise}
\end{cases}
\end{equation}
where valid outputs must contain exactly one \texttt{<think>}...\texttt{</think>} reasoning block followed by a \texttt{<solution>}...\texttt{</solution>} block, $J(Y, \hat{Y}) = |Y \cap \hat{Y}| / |Y \cup \hat{Y}|$ is the Jaccard similarity score between predicted and ground truth labels, and $\lambda_{\text{len}}$ is a penalty coefficient for short responses.

The \emph{nuanced reward} implements a multi-component scoring mechanism designed to provide a more granular reward signal during training:
\begin{equation}
r_{\text{nuanced}}(t, Y) = r_{\text{match}} + r_{\text{partial}} - r_{\text{FP}} - r_{\text{collapse}} - r_{\text{format}},
\end{equation}
where exact matches receive +100 points, partial credit is scaled by recall (30 points per correct label relative to ground truth size) and precision (20 points per correct label relative to prediction size), and various penalties discourage undesirable behaviors. Critically, we incorporated frequency-weighted penalties for false positives: commonly occurring labels such as ``No Finding'' and ``Support Devices'' incur higher penalties when incorrectly predicted than rare pathologies, discouraging shotgun predictions. To prevent mode collapse, we monitor a sliding window of the most recent 100 predictions and apply cumulative penalties ($-30$ per excess repetition, $-50$ for mode collapse) when any single label exceeds 70\% dominance. Additional penalties discourage invalid CheXpert labels ($-100$ each), duplicate predictions ($-25$ each), and extraneous text outside the designated blocks. All predicted labels are filtered to the 14 valid CheXpert labels before evaluation.

\section{Results and Evaluation}

\subsection{Effect of Prompting}
To identify the optimal instruction format for multilabel CheXpert classification, we evaluated nine prompt variants on MedGemma-4B, selected as the primary ablation target due to its superior base performance among available models. Each variant implemented a distinct reasoning strategy, ranging from structured, stepwise protocols to free-form narratives and explicit verification mechanisms (Table~\ref{tab:prompt_ablation}). We initially prioritized a structured approach, denoted as \textit{Reasoning A}, \
which aimed to mimic radiological best practices developed by experts from the \
University Clinic Bonn and Queensland Health. This prompt instructed the LLM to perform a mandatory 12-step chest X-ray analysis 
within \texttt{<think>} tags, followed by a strict list of derived CheXpert labels 
within \texttt{<solution>} tags. Contrary to expectations, the less constrained \textit{Reasoning Narrative} prompt 
achieved the highest overall performance (micro-F1 = $0.524$, macro-F1 = $0.270$), 
followed by \textit{Reasoning Self-Check} (micro-F1 = $0.514$, macro-F1 = $0.253$). 
Both outperformed the structured \textit{Reasoning A} baseline 
(micro-F1 = $0.498$, macro-F1 = $0.245$). The performance differential was driven primarily by recall rather than precision. 
\textit{Reasoning Narrative} and \textit{Reasoning Self-Check} attained 
substantially higher overall recall ($0.599$ and $0.558$, respectively) 
compared to \textit{Reasoning A} ($0.504$), suggesting that these prompt framings 
enhanced the model's sensitivity to positive findings. Conversely, certain complex prompts introduced significant decoding failures, most 
notably \textit{Reasoning C} (failure rate $0.482$) and \textit{Reasoning F} ($0.180$), which 
severely hindered end-to-end performance by producing unparseable outputs. Prompts that imposed overly rigid formatting constraints (Reasoning C, Reasoning F) or required explicit differential comparisons for every finding (Reasoning F) suffered from both decoding failures and reduced classification quality. In contrast, free-form narrative reasoning (Reasoning Narrative) and structured self-checking (Reasoning Self-Check) balanced interpretability with output reliability, yielding micro-F1 gains of +0.026 and +0.016 over the baseline. Notably, these results reflect MedGemma-4B's medical pre-training; as we show in the SFT analysis, the structured \textit{Reasoning A} format, which underperformed here, becomes the optimal choice for general-purpose models lacking domain-specific knowledge.

\begin{table}[H]
\centering
\caption{Prompt ablation on base MedGemma-4B (CheXpert validation).}
\label{tab:prompt_ablation}
\setlength{\tabcolsep}{12pt}
\begin{tabular}{@{}lrrr@{}}
	\toprule
	\textbf{Prompt Configuration} & \multicolumn{3}{c}{\textbf{CheXpert Validation Metrics}} \\
\cmidrule(r){2-4}
 & Micro-F1 & Macro-F1 & Fail Rate \\
\midrule
	\textbf{Reasoning Narrative} & \textbf{0.524} & \textbf{0.270} & 0.002 \\
Reasoning Self-Check & 0.514 & 0.253 & 0.010 \\
Reasoning Hypothesis & 0.357 & 0.188 & 0.022 \\
Reasoning A (baseline) & 0.498 & 0.245 & 0.000 \\
Reasoning B & 0.427 & 0.170 & 0.000 \\
Reasoning C & 0.260 & 0.132 & 0.482 \\
Reasoning D & 0.329 & 0.170 & 0.000 \\
Reasoning E & 0.404 & 0.218 & 0.000 \\
Reasoning F & 0.269 & 0.108 & 0.180 \\
\midrule
\multicolumn{4}{l}{\footnotesize\textit{Key design cues:}} \\
\multicolumn{4}{l}{\footnotesize \textbf{Reasoning Narrative}: Free-form expert narrative with evidence aggregation} \\
\multicolumn{4}{l}{\footnotesize \quad Reasoning Self-Check: Region-wise scan then explicit verification pass} \\
\multicolumn{4}{l}{\footnotesize \quad Reasoning Hypothesis: Hypothesis-driven comparison of candidate findings} \\
\multicolumn{4}{l}{\footnotesize \quad Reasoning A: Structured 12-step checklist with strict format (Baseline)} \\
\multicolumn{4}{l}{\footnotesize \quad Reasoning B: Zonal lung review with pathology differentiation} \\
\multicolumn{4}{l}{\footnotesize \quad Reasoning C: Evidence-heavy report with label-confidence matrix} \\
\multicolumn{4}{l}{\footnotesize \quad Reasoning D: Definition-guided labeling with strict criteria} \\
\multicolumn{4}{l}{\footnotesize \quad Reasoning E: Second-look pass over anatomic blind spots} \\
\multicolumn{4}{l}{\footnotesize \quad Reasoning F: Differential diagnosis required for each abnormality} \\
\bottomrule
\end{tabular}
\end{table}

\subsection{Effect of Supervised Fine-Tuning}

To investigate whether instruction format effectiveness depends on medical pre-training, we fine-tuned two vision-language models, MedGemma-4B (medically pre-trained) and Qwen2.5-VL-3B-Instruct (general-purpose, no medical training), on four distinct instruction formats: \textit{Reasoning Narrative} (free-form radiologic narrative), \textit{Reasoning A} (structured 12-step reasoning), \textit{Free Reasoning} (concise analysis followed by labels), and \textit{Only Label} (direct label output with no rationale).
\bigskip
\bigskip

\textbf{MedGemma-4B Results.} For the medically pre-trained model, a striking trade-off emerged between micro-F1 and macro-F1 performance (Table~\ref{tab:sft_comparison}). The \textit{Only Label} variant achieved the highest micro-F1 (0.461), while \textit{Free Reasoning} obtained the best macro-F1 (0.253). This divergence reflects label frequency effects: direct label prediction excels at high-support conditions like ``No Finding'' and ``Support Devices'' that dominate micro-averaged metrics, but struggles with rare pathologies that carry equal weight in macro-averages. Notably, \textit{Reasoning A} underperformed significantly across both metrics (micro-F1 = 0.293, macro-F1 = 0.139), suggesting that explicit structured reasoning provides little benefit when the model already possesses domain-specific feature representations.

\textbf{Qwen2.5-VL-3B-Instruct Results.} The general-purpose model exhibited a strikingly different pattern (Table~\ref{tab:sft_comparison_qwen}). \textit{Reasoning A}, the structured 12-step reasoning format, achieved the best performance by a substantial margin (micro-F1 = 0.371, macro-F1 = 0.208), outperforming all other variants. In contrast, \textit{Only Label}, which succeeded on MedGemma, dropped to mediocre performance (micro-F1 = 0.249, macro-F1 = 0.080). \textit{Free Reasoning} also struggled (micro-F1 = 0.200, macro-F1 = 0.131), while \textit{Reasoning Narrative} showed moderate scores but high failure rates (17\%).

\textbf{Cross-Model Interpretation.} This complete ranking reversal reveals a fundamental insight: medical pre-training encodes domain-specific feature representations that enable direct label mapping, whereas general-purpose VLMs require explicit structured reasoning scaffolds to compensate for missing domain knowledge. The 12-step clinical analysis in \textit{Reasoning A} provides necessary interpretive guidance for Qwen, directing attention through systematic examination of anatomical structures and pathological indicators. However, this same structure appears redundant for MedGemma, which has already internalized these reasoning patterns through medical pre-training. The failure of free-form reasoning on Qwen (but moderate success on MedGemma) further supports this hypothesis: without either pre-trained medical representations or explicit structural guidance, the model lacks the necessary inductive biases for clinical interpretation.

\begin{figure}[H]
\centering
\includegraphics[width=\linewidth]{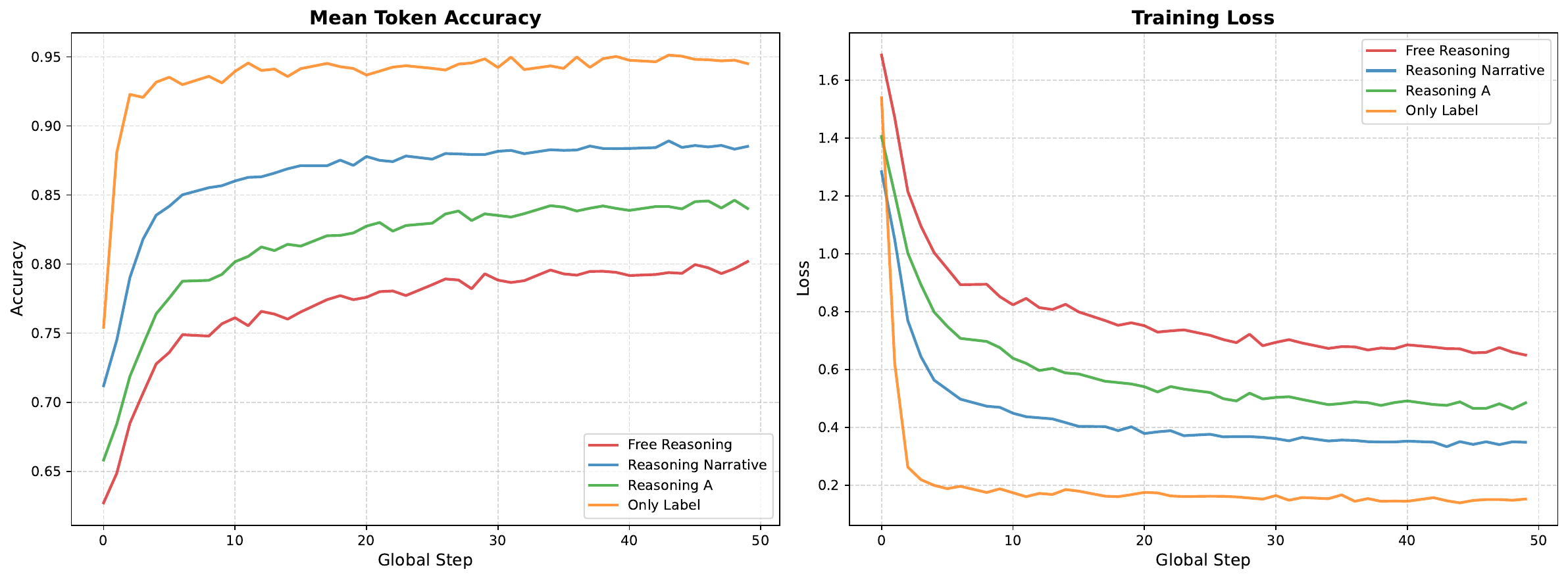}
\caption{Comparative training dynamics across all four supervised fine-tuning variants. The right panel shows training loss convergence, while the left panel displays mean token-level prediction accuracy. Notably, syntax-constrained variants (Only Label, Reasoning A, Reasoning Narrative) converge rapidly to higher token accuracy saturation, whereas Free Reasoning exhibits slower convergence and lower saturation levels. This pattern reflects fundamental differences in output entropy rather than learning quality, with Free Reasoning's diverse, unstructured traces requiring more nuanced semantic learning compared to the predictable template patterns of structured formats.}
\label{fig:sft_training_metrics}
\end{figure}

\textbf{Decoding Reliability and Training Dynamics.} Across both models, decoding reliability correlated inversely with rationale length, with \textit{Only Label} producing zero format violations while longer formats increased malformation risk. The training dynamics in Figure~\ref{fig:sft_training_metrics} show that syntax-constrained variants (Only Label, Reasoning A) converge rapidly to high token accuracy, while Free Reasoning converges gradually and saturates lower. We hypothesize this reflects output entropy rather than learning quality: templated formats have predictable token distributions that inflate accuracy metrics but encourage pattern memorization, whereas free-form traces force the model to learn semantic relationships rather than surface shortcuts.

Given its superior macro-F1 performance and balanced handling of both common and rare pathologies on the medically pre-trained model, we selected the MedGemma-4B \textit{Free Reasoning} variant as the initialization checkpoint for subsequent GRPO training.

\begin{table}[H]
\centering
\caption{Supervised fine-tuning variants on MedGemma-4B (CheXpert validation).}
\label{tab:sft_comparison}
\resizebox{\linewidth}{!}{%
\begin{tabular}{lccc ccc c}
 & \multicolumn{3}{c}{Micro} & \multicolumn{3}{c}{Macro} & \multirow{2}{*}{Fail} \\
\cmidrule(lr){2-4} \cmidrule(lr){5-7}
Variant & P & R & F1 & P & R & F1 & \\
\midrule
Only Label & 0.596 & 0.375 & \textbf{0.461} & 0.332 & 0.214 & 0.241 & 0.000 \\
Free Reasoning & 0.480 & 0.365 & 0.415 & 0.358 & 0.227 & \textbf{0.253} & 0.032 \\
Reasoning Narrative & 0.565 & 0.281 & 0.376 & 0.345 & 0.141 & 0.180 & 0.210 \\
Reasoning A & 0.405 & 0.230 & 0.293 & 0.381 & 0.143 & 0.139 & 0.004 \\
\bottomrule
\end{tabular}%
}
\end{table}

	\begin{table}[H]
	\centering
	\caption{Supervised fine-tuning variants on Qwen/Qwen2.5-VL-3B-Instruct (CheXpert validation).}
	\label{tab:sft_comparison_qwen}
	\resizebox{\linewidth}{!}{%
	\begin{tabular}{lccc ccc c}
	 & \multicolumn{3}{c}{Micro} & \multicolumn{3}{c}{Macro} & \multirow{2}{*}{Fail} \\
	\cmidrule(lr){2-4} \cmidrule(lr){5-7}
	Variant & P & R & F1 & P & R & F1 & \\
	\midrule
	Only Label & 0.333 & 0.199 & 0.249 & 0.153 & 0.122 & 0.080 & 0.000 \\
	Free Reasoning & 0.200 & 0.200 & 0.200 & 0.150 & 0.142 & 0.131 & 0.056 \\
	Reasoning Narrative & 0.396 & 0.198 & 0.264 & 0.122 & 0.076 & 0.075 & 0.170 \\
	Reasoning A & 0.322 & 0.436 & \textbf{0.371} & 0.194 & 0.265 & \textbf{0.208} & 0.002 \\
	\bottomrule
	\end{tabular}%
	}
	\end{table}
	
\subsection{Effect of Group Relative Policy Optimization}

We evaluated GRPO training using both reward functions on the \textit{Free Reasoning} SFT checkpoint (Table~\ref{tab:rl_comparison}). 
The simpler \textit{hard reward} function achieved marginally better validation performance (micro-F1 = 0.391, macro-F1 = 0.258) 
compared to the complex \textit{nuanced reward} (micro-F1 = 0.387, macro-F1 = 0.257), suggesting that explicit format 
enforcement combined with Jaccard similarity provided sufficient signal for policy optimization. 
Notably, both reward functions improved macro-F1 relative to the SFT checkpoint (0.253 → 0.258), 
indicating enhanced performance on low-prevalence, diagnostically challenging conditions, 
though micro-F1 declined slightly (0.415 → 0.391), reflecting modest performance degradation on 
high-support labels that dominate the micro-averaged metric.

\begin{figure}[H]
	\centering
	\includegraphics[width=\linewidth]{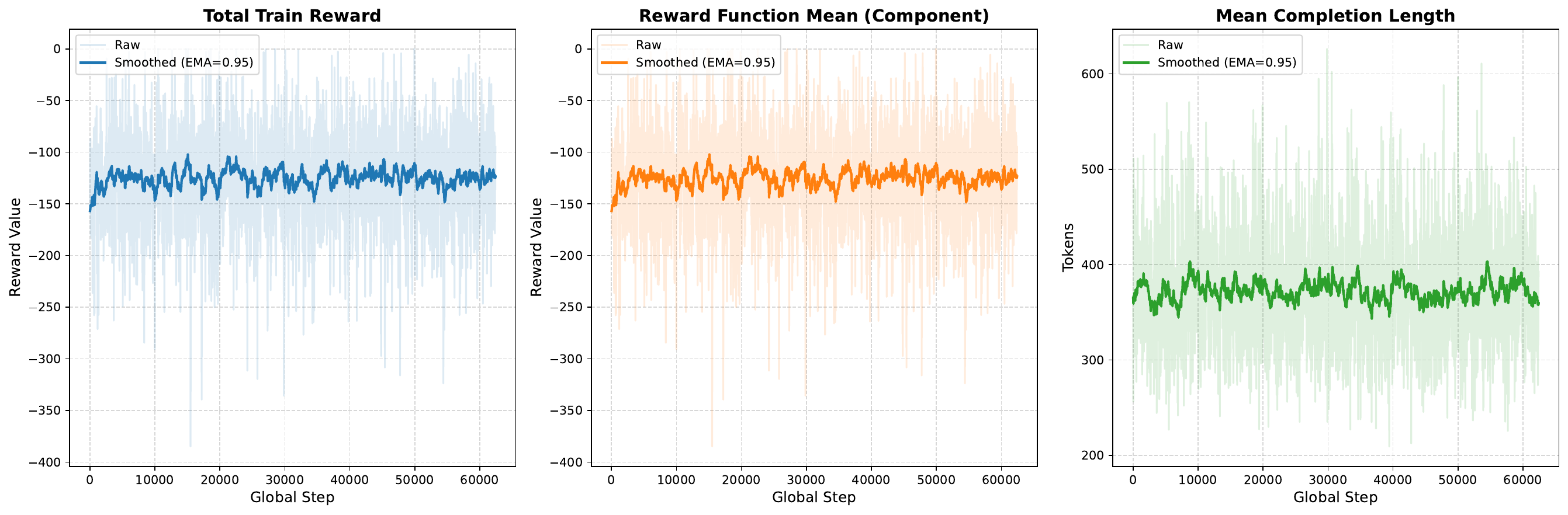}
	\caption{Evolution of training metrics during the GRPO training process using \textit{nuanced rewards}. The panels display (left) the total training reward, (center) the mean reward function component, and (right) the mean completion length in tokens. Faint lines represent raw data recorded at each global step, while bold lines indicate an exponential moving average (EMA) with a smoothing factor of 0.95 to highlight the underlying training trends.}
	\label{fig:output}
	\end{figure}
	
	\begin{figure}[H]
	\centering
	\includegraphics[width=\linewidth]{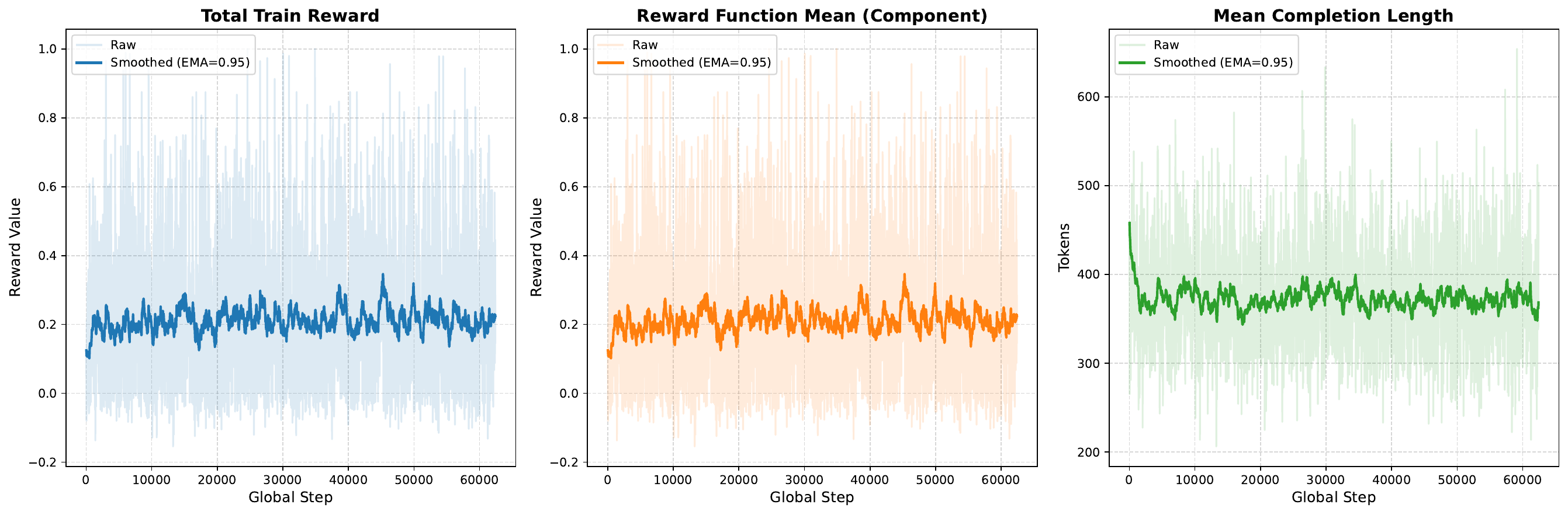}
	\caption{Evolution of training metrics during the GRPO training process using \textit{hard rewards}. The panels display (left) the total training reward, (center) the mean reward function component, and (right) the mean completion length in tokens. Faint lines represent raw data recorded at each global step, while bold lines indicate an exponential moving average (EMA) with a smoothing factor of 0.95 to highlight the underlying training trends.}
	\label{fig:output_new}
	\end{figure}

On the in-distribution CheXpert test set (Table~\ref{tab:comparative_category_f1_merged_extended}), 
GRPO training successfully recovered the performance lost during supervised fine-tuning. 
The ChexReason model (SFT + GRPO) achieved a macro-F1 of 0.346, representing a 23\% improvement over the 
SFT checkpoint (0.282) and nearly matching the MedGemma baseline (0.362). 
This recovery was particularly pronounced for categories such as Cardiomegaly (0.442 → 0.664), 
Lung Opacity (0.161 → 0.743), and Support Devices (0.728 → 0.818), where the reward signal 
effectively guided the model toward more accurate label predictions.

The generalization characteristics of GRPO training reveal a more nuanced picture when evaluated across 
two out-of-distribution test sets. On the CheXpert test set (Table~\ref{tab:comparative_category_f1_merged_extended}), which
originates from Stanford Hospital but shares the same 14-label CheXpert schema as the MIMIC-CXR-JPG training
data, the ChexReason model achieves strong performance (macro-F1 = 0.346), substantially improving over the SFT 
checkpoint (0.282 → 0.346, +23\%). However, on the NIH Chest X-ray test set (Table~\ref{tab:comparative_category_f1_merged}), which
employs a distinct labeling methodology reduced to nine CheXpert-compatible categories, the ChexReason model regresses
to baseline levels (macro-F1 = 0.243), representing a 19\% degradation relative to the SFT checkpoint 
(0.299 → 0.243) and matching the pretrained MedGemma baseline performance.

This divergent behavior across out-of-distribution datasets suggests that GRPO optimization aligns the model 
to the semantic structure of the CheXpert labeling convention rather than to generalizable radiologic features. 
The shared label schema between MIMIC-CXR-JPG (training) and CheXpert (test), encompassing identical pathology
definitions, labeling granularity, and uncertainty encoding, enables the reward signal to exploit label-specific 
patterns that transfer within this ecosystem but fail to generalize to alternative labeling methodologies. 
Paradoxically, the SFT checkpoint achieves its best performance on the NIH dataset (macro-F1 = 0.299) despite 
exhibiting the weakest scores on the CheXpert test set (0.282), suggesting that teacher-generated reasoning 
traces capture visual-semantic relationships that transcend specific label taxonomies, even as they degrade 
performance on the validation distribution used to guide training.

These findings reveal a fundamental tension in reinforcement learning for small vision-language models in 
medical imaging: reward-based optimization excels at recovering performance within a specific labeling framework 
but appears to fail eliciting the kind of generalizable diagnostic reasoning that transfers across institutional conventions 
and annotation methodologies. The fact that the simpler hard reward performs comparably to the carefully engineered 
nuanced reward further supports this interpretation. Both reward functions provide sufficient supervision to align the model to 
CheXpert label semantics, but neither addresses the underlying challenge of learning institution-agnostic radiologic 
representations. This suggests that improving cross-dataset generalization may require architectural modifications, 
multi-dataset training curricula, or reward formulations that explicitly penalize overfitting to labeling conventions 
rather than post-hoc RL fine-tuning on single-institution data.

\begin{table}[H]
	\centering
	\caption{Reinforcement learning using GRPO in various setups (CheXpert validation). All variants use the \textbf{MedGemma-4B Free-Reasoning SFT} model checkpoint.}
	\label{tab:rl_comparison}
	\setlength{\tabcolsep}{5pt} 
	\begin{tabular}{@{}ll ccc ccc c@{}}
	\toprule
	& & \multicolumn{3}{c}{\textbf{Micro}} & \multicolumn{3}{c}{\textbf{Macro}} & \\
	\cmidrule(lr){3-5} \cmidrule(lr){6-8}
	\textbf{\shortstack[l]{Reward\\Function}} & \textbf{Training Steps} & P & R & F1 & P & R & F1 & \textbf{Fail} \\
	\midrule
	\multicolumn{9}{@{}l}{\textit{Nuanced}} \\
	& Best F1-Score & 0.342 & 0.445 & 0.387 & 0.248 & 0.311 & 0.257 & 0.020 \\
	& Maximum  & 0.335 & 0.424 & 0.374 & 0.228 & 0.271 & 0.232 & 0.030 \\
	\addlinespace
	\multicolumn{9}{@{}l}{\textit{Hard}} \\
	& Best F1-Score & \textbf{0.343} & \textbf{0.455} & \textbf{0.391} & \textbf{0.250} & \textbf{0.328} & \textbf{0.258} & 0.028 \\
	& Maximum & 0.340 & 0.418 & 0.375 & 0.243 & 0.279 & 0.242 & \textbf{0.018} \\
	\bottomrule
	\end{tabular}
	\end{table}

\subsection{Generalization Analysis}

The evolution of SFT training metrics in Figure~\ref{fig:sft_training_metrics} indicates that the \textit{Free Reasoning} variant successfully masters the structured output format with stable convergence. However, this optimization masks complex generalization challenges revealed during out-of-distribution testing.

A striking pattern emerges in cross-dataset transfer. High-performing models like NV-Reason (macro-F1 = 0.755 on CheXpert) and ChexReason (0.346) exhibit dramatic degradation on the NIH dataset, dropping 61\% and 30\% respectively (Tables~\ref{tab:comparative_category_f1_merged_extended} and \ref{tab:comparative_category_f1_merged}). This parallel failure suggests aggressive optimization for CheXpert-specific patterns trades generalization for benchmark performance. Empirical evidence supports this: Compton et al.~\cite{compton2023} demonstrated that models often rely on deep-seated dataset artifacts like hospital source discrimination. In contrast, the SFT checkpoint uniquely improves on NIH (0.282 $\rightarrow$ 0.299) despite weaker CheXpert scores, whereas the MedGemma baseline drops 33\%.

We hypothesize this stems from the susceptibility of small vision-language models (3-4B parameters) to spurious patterns. Murali et al.~\cite{murali2023} demonstrated that spurious features are learned early and preferentially when they contain high ``usable information,'' a process entrenched by subsequent optimization on benchmark datasets. However, our teacher-guided SFT process appears to disrupt this overfitting without re-introducing benchmark-specific pressures.

Aligned with knowledge distillation literature, Boland et al.~\cite{boland2025} showed that distillation from unbiased teachers significantly reduces spurious feature learning. By forcing alignment with Gemini-generated reasoning traces that emphasize diagnostic principles rather than shortcuts, our SFT approach acts as implicit knowledge distillation. These relearned, institution-agnostic principles prove less effective on the artifact-heavy CheXpert distribution but transfer more reliably to the NIH dataset.

The observation that NV-Reason and ChexReason lose this generalization advantage when optimized for higher CheXpert scores supports the notion that current benchmark-driven development practices may inadvertently disfavor models with broader real-world viability. This interpretation aligns with Vogt-Lowell et al.~\cite{vogtlowell2023}, who report that end-to-end fine-tuning can compromise out-of-distribution robustness. Collectively, these findings imply that the deliberate constraint of supervised fine-tuning with reasoning-aligned guidance may help preserve institution-agnostic generalization capabilities by disrupting the learned shortcuts often associated with competitive benchmark performance.

\begin{table}[H]
\centering
    \caption{Comparative F1-Score performance across five models on the out-of-sample original Chexpert dataset, including overall aggregated performance. The best performing model for each category is highlighted in bold.}
    \label{tab:comparative_category_f1_merged_extended}
    \resizebox{\linewidth}{!}{%
    \begin{tabular}{lccccc}
        \toprule
     & \multicolumn{5}{c}{\textbf{F1-Score}} \\
    \cmidrule(lr){2-6}
        \textbf{Category} & \textbf{NV-Reason} & \textbf{MedGemma} & \textbf{Qwen2.5} & \multicolumn{2}{c}{\textbf{MedGemma}} \\
    \cmidrule(lr){5-6}
         & & & & \textbf{(SFT)} & \textbf{(ChexReason)} \\
    \midrule
    Atelectasis & \textbf{0.758} & 0.288 & 0.354 & 0.229 & 0.240 \\
    Cardiomegaly & \textbf{0.853} & 0.627 & 0.385 & 0.442 & 0.664 \\
    Consolidation & \textbf{0.188} & 0.000 & 0.097 & \textbf{0.205} & 0.000 \\
    Edema & \textbf{0.807} & 0.174 & 0.128 & 0.385 & 0.132 \\
    Enlarged Cardiomediastinum & \textbf{0.879} & 0.016 & 0.407 & 0.075 & 0.000 \\
    Fracture & \textbf{0.750} & 0.143 & 0.009 & 0.150 & 0.200 \\
    Lung Lesion & \textbf{0.875} & 0.300 & 0.042 & 0.211 & 0.133 \\
    Lung Opacity & \textbf{0.961} & 0.748 & 0.464 & 0.161 & 0.743 \\
    No Finding & \textbf{0.775} & 0.625 & 0.318 & 0.579 & 0.607 \\
    Pleural Effusion & \textbf{0.822} & 0.634 & 0.264 & 0.464 & 0.625 \\
    Pleural Other & \textbf{0.667} & 0.000 & 0.026 & 0.000 & 0.000 \\
    Pneumonia & \textbf{0.429} & 0.400 & 0.031 & 0.143 & 0.400 \\
    Pneumothorax & \textbf{0.842} & 0.308 & 0.033 & 0.179 & 0.286 \\
    Support Devices & \textbf{0.970} & 0.808 & 0.637 & 0.728 & 0.818 \\
    \midrule
        \textit{Overall Average (Macro F1)} & \textbf{0.755} & 0.362 & 0.228 & 0.282 & 0.346 \\
    \bottomrule
    \end{tabular}
    }
\end{table}

\begin{table}[H]
\centering
\caption{Comparative F1-Score performance across five models on the out-of-distribution dataset, including overall aggregated performance. The best performing model for each category is highlighted in bold.}
\label{tab:comparative_category_f1_merged}
\resizebox{\linewidth}{!}{%
\begin{tabular}{lccccc}
	\toprule
 & \multicolumn{5}{c}{\textbf{F1-Score}} \\
\cmidrule(lr){2-6}
	\textbf{Category} & \textbf{NV-Reason} & \textbf{MedGemma} & \textbf{Qwen2.5} & \multicolumn{2}{c}{\textbf{MedGemma}} \\
\cmidrule(lr){5-6}
	 & & & & \textbf{(SFT)} & \textbf{(ChexReason)} \\
\midrule
Atelectasis & \textbf{0.422} & 0.300 & 0.000 & 0.264 & 0.260 \\
Cardiomegaly & \textbf{0.543} & 0.482 & 0.000 & 0.440 & 0.312 \\
Consolidation & 0.056 & 0.029 & 0.109 & \textbf{0.200} & 0.178 \\
Edema & 0.273 & 0.522 & 0.000 & \textbf{0.526} & 0.429 \\
Lung Lesion & \textbf{0.397} & 0.195 & 0.115 & 0.113 & 0.179 \\
No Finding & 0.283 & 0.275 & 0.215 & \textbf{0.343} & 0.250 \\
Pleural Other & 0.152 & 0.180 & 0.000 & \textbf{0.248} & 0.082 \\
Pneumonia & 0.000 & \textbf{0.202} & 0.117 & 0.138 & 0.182 \\
Pneumothorax & \textbf{0.552} & 0.000 & 0.000 & 0.416 & 0.317 \\
\midrule
	\textit{Overall Average (Macro F1)} & 0.297 & 0.243 & 0.062 & \textbf{0.299} & 0.243 \\
\bottomrule
\end{tabular}
}
\end{table}

\section{Conclusion}

This study examined whether R1-style training, combining supervised fine-tuning with Group Relative Policy Optimization, enhances multilabel chest X-ray classification in small vision-language models under low-resource conditions. We trained MedGemma-4B and Qwen2.5-VL-3B-Instruct using teacher-generated reasoning traces from Gemini 2.5, employing only 2,000 SFT and 1,000 RL samples on a single A100 GPU. Our ChexReason model achieved a 23\% improvement over the SFT checkpoint on the standard CheXpert benchmark (macro-F1 = 0.346), despite training on out-of-distribution MIMIC-CXR-JPG data.

However, this success severely compromised cross-dataset generalization. On the NIH Chest X-ray dataset, which uses different labeling methodology, ChexReason performance degraded by 19\%, reverting to baseline levels (macro-F1 = 0.243). This mirrors the high-resource NV-Reason-CXR-3B model, which also saw massive drops on NIH data despite state-of-the-art CheXpert scores. Paradoxically, our SFT checkpoint uniquely improved on NIH (0.282 $\rightarrow$ 0.299 macro-F1) while showing weaker CheXpert performance, suggesting teacher-guided reasoning traces capture more generalizable visual-semantic relationships than reward-optimized outputs. Our cross-model comparison further revealed that instruction format effectiveness depends on medical pre-training. Qwen2.5-VL-3B, lacking domain-specific representations, performed best with explicit 12-step structured reasoning (macro-F1 = 0.208), while MedGemma-4B performed best with direct label prediction (macro-F1 = 0.253). This suggests structured reasoning scaffolds can compensate for missing domain knowledge, but become redundant, or even detrimental, when medical pre-training has already internalized clinical reasoning patterns. These findings reveal a tension in reinforcement learning for small medical VLMs: reward signals recover benchmark performance by exploiting dataset-specific semantics, but potentially degrade transferability across institutions. Since both the high-resource NVIDIA model and our low-resource ChexReason model failed similarly, the issue likely stems from the RL fine-tuning paradigm itself when applied to small models on standardized benchmarks. Consequently, R1-style training on hard labels such as CheXpert may be counterproductive for clinical deployment requiring robustness across diverse populations. Practitioners under resource constraints may be better served by curated, supervised fine-tuning rather than aggressive benchmark optimization.

\bibliographystyle{plain} 
\bibliography{Distillation}

@misc{johnson_mimic-cxr-jpg_2019,
	title = {{MIMIC}-{CXR}-{JPG}, a large publicly available database of labeled chest radiographs},
	url = {https://arxiv.org/abs/1901.07042},
	author = {Johnson, Alistair E. W. and others},
	year = {2019},
	note = {arXiv:1901.07042},
}

@article{okolo_cln_2025,
	title = {{CLN}: {A} multi-task deep neural network for chest {X}-ray image localisation and classification},
	volume = {288},
	issn = {09574174},
	shorttitle = {{CLN}},
	url = {https://linkinghub.elsevier.com/retrieve/pii/S0957417425017828},
	doi = {10.1016/j.eswa.2025.128162},
	language = {en},
	urldate = {2025-10-07},
	journal = {Expert Systems with Applications},
	author = {Okolo, Gabriel Iluebe and Katsigiannis, Stamos and Ramzan, Naeem},
	month = sep,
	year = {2025},
	pages = {128162},
}

@article{lotfinia_boosting_2025,
	title = {Boosting multi-demographic federated learning for chest radiograph analysis using general-purpose self-supervised representations},
	volume = {3},
	issn = {30505771},
	url = {https://linkinghub.elsevier.com/retrieve/pii/S305057712500026X},
	doi = {10.1016/j.ejrai.2025.100028},
	language = {en},
	urldate = {2025-10-07},
	journal = {European Journal of Radiology Artificial Intelligence},
	author = {Lotfinia, Mahshad and Tayebiarasteh, Arash and Samiei, Samaneh and Joodaki, Mehdi and Tayebi Arasteh, Soroosh},
	month = sep,
	year = {2025},
	pages = {100028},
}

@article{abdullah_automated_2025,
	title = {Automated {Radiology} {Report} {Labeling} in {Chest} {X}-{Ray} {Pathologies}: {Development} and {Evaluation} of a {Large} {Language} {Model} {Framework}},
	volume = {13},
	issn = {2291-9694},
	shorttitle = {Automated {Radiology} {Report} {Labeling} in {Chest} {X}-{Ray} {Pathologies}},
	url = {https://medinform.jmir.org/2025/1/e68618},
	doi = {10.2196/68618},
	urldate = {2025-10-07},
	journal = {JMIR Medical Informatics},
	author = {Abdullah, Abdullah and Kim, Seong Tae},
	month = mar,
	year = {2025},
	pages = {e68618--e68618},
}

@misc{zhang_rexrank_2024,
	title = {{ReXrank}: {A} {Public} {Leaderboard} for {AI}-{Powered} {Radiology} {Report} {Generation}},
	shorttitle = {{ReXrank}},
	url = {http://arxiv.org/abs/2411.15122},
	doi = {10.48550/arXiv.2411.15122},
	urldate = {2025-10-07},
	publisher = {arXiv},
	author = {Zhang, Xiaoman and Zhou, Hong-Yu and Yang, Xiaoli and Banerjee, Oishi and Acosta, Julián N. and Miller, Josh and Huang, Ouwen and Rajpurkar, Pranav},
	month = nov,
	year = {2024},
	note = {arXiv:2411.15122 [cs]},
}

@article{tanno_collaboration_2025,
	title = {Collaboration between clinicians and vision–language models in radiology report generation},
	volume = {31},
	issn = {1078-8956, 1546-170X},
	url = {https://www.nature.com/articles/s41591-024-03302-1},
	doi = {10.1038/s41591-024-03302-1},
	language = {en},
	number = {2},
	urldate = {2025-10-07},
	journal = {Nature Medicine},
	author = {Tanno, Ryutaro and Barrett, David G. T. and Sellergren, Andrew and Ghaisas, Sumedh and Dathathri, Sumanth and See, Abigail and Welbl, Johannes and Lau, Charles and Tu, Tao and Azizi, Shekoofeh and Singhal, Karan and Schaekermann, Mike and May, Rhys and Lee, Roy and Man, SiWai and Mahdavi, Sara and Ahmed, Zahra and Matias, Yossi and Barral, Joelle and Eslami, S. M. Ali and Belgrave, Danielle and Liu, Yun and Kalidindi, Sreenivasa Raju and Shetty, Shravya and Natarajan, Vivek and Kohli, Pushmeet and Huang, Po-Sen and Karthikesalingam, Alan and Ktena, Ira},
	month = feb,
	year = {2025},
	pages = {599--608},
}

@misc{bhardwaj_enhancing_2025,
	title = {Enhancing zero-shot learning in medical imaging: integrating clip with advanced techniques for improved chest x-ray analysis},
	shorttitle = {Enhancing zero-shot learning in medical imaging},
	url = {http://arxiv.org/abs/2503.13134},
	doi = {10.48550/arXiv.2503.13134},
	urldate = {2025-10-07},
	publisher = {arXiv},
	author = {Bhardwaj, Prakhar and Bhat, Sheethal and Maier, Andreas},
	month = mar,
	year = {2025},
	note = {arXiv:2503.13134 [cs]},
}

@misc{myronenko_nv-reason-cxr_2025,
    title = {Reasoning Visual Language Model for Chest X-Ray Analysis},
    author = {Myronenko, Andriy and Yang, Dong and Turkbey, Baris and Salama, Paul and others},
    year = {2025},
    doi = {10.48550/arXiv.2510.23968},
    url = {https://arxiv.org/abs/2510.23968},
    note = {arXiv:2510.23968 [cs.CV]},
}

@article{jang_significantly_2024,
	title = {Significantly improving zero-shot {X}-ray pathology classification via fine-tuning pre-trained image-text encoders},
	volume = {14},
	issn = {2045-2322},
	url = {https://www.nature.com/articles/s41598-024-73695-z},
	doi = {10.1038/s41598-024-73695-z},
	language = {en},
	number = {1},
	urldate = {2025-10-07},
	journal = {Scientific Reports},
	author = {Jang, Jongseong and Kyung, Daeun and Kim, Seung Hwan and Lee, Honglak and Bae, Kyunghoon and Choi, Edward},
	month = oct,
	year = {2024},
	pages = {23199},
}

@misc{wang_cxpmrg-bench_2024,
	title = {{CXPMRG}-{Bench}: {Pre}-training and {Benchmarking} for {X}-ray {Medical} {Report} {Generation} on {CheXpert} {Plus} {Dataset}},
	shorttitle = {{CXPMRG}-{Bench}},
	url = {http://arxiv.org/abs/2410.00379},
	doi = {10.48550/arXiv.2410.00379},
	urldate = {2025-10-07},
	publisher = {arXiv},
	author = {Wang, Xiao and Wang, Fuling and Li, Yuehang and Ma, Qingchuan and Wang, Shiao and Jiang, Bo and Li, Chuanfu and Tang, Jin},
	month = oct,
	year = {2024},
	note = {arXiv:2410.00379 [cs]},
}

@misc{ng_x-ray-cot_2025,
	title = {X-{Ray}-{CoT}: {Interpretable} {Chest} {X}-ray {Diagnosis} with {Vision}-{Language} {Models} via {Chain}-of-{Thought} {Reasoning}},
	shorttitle = {X-{Ray}-{CoT}},
	url = {http://arxiv.org/abs/2508.12455},
	doi = {10.48550/arXiv.2508.12455},
	urldate = {2025-10-17},
	publisher = {arXiv},
	author = {Ng, Chee and Sun, Liliang and Tang, Shaoqing},
	month = aug,
	year = {2025},
	note = {arXiv:2508.12455 [cs]},
}

@misc{fan_chestx-reasoner_2025,
	title = {{ChestX}-{Reasoner}: {Advancing} {Radiology} {Foundation} {Models} with {Reasoning} through {Step}-by-{Step} {Verification}},
	shorttitle = {{ChestX}-{Reasoner}},
	url = {http://arxiv.org/abs/2504.20930},
	doi = {10.48550/arXiv.2504.20930},
	urldate = {2025-10-17},
	publisher = {arXiv},
	author = {Fan, Ziqing and Liang, Cheng and Wu, Chaoyi and Zhang, Ya and Wang, Yanfeng and Xie, Weidi},
	month = may,
	year = {2025},
	note = {arXiv:2504.20930 [cs]},
}

@misc{fallahpour_medrax_2025,
	title = {{MedRAX}: {Medical} {Reasoning} {Agent} for {Chest} {X}-ray},
	shorttitle = {{MedRAX}},
	url = {http://arxiv.org/abs/2502.02673},
	doi = {10.48550/arXiv.2502.02673},
	urldate = {2025-10-17},
	publisher = {arXiv},
	author = {Fallahpour, Adibvafa and Ma, Jun and Munim, Alif and Lyu, Hongwei and Wang, Bo},
	month = may,
	year = {2025},
	note = {arXiv:2502.02673 [cs]},
}

@misc{hu_medical-cxr-vqa_2025,
	title = {{Medical-CXR-VQA} dataset: {A} {Large-Scale} {LLM-Enhanced} {Medical} {Dataset} for {Visual} {Question} {Answering} on {Chest} {X-Ray} {Images}},
	author = {Hu, X. and Gu, L. and Kobayashi, K. and Liu, L. and Zhang, M. and Harada, T. and Summers, R. and Zhu, Y.},
	year = {2025},
	publisher = {PhysioNet},
	doi = {10.13026/1pm5-hy02},
	url = {https://doi.org/10.13026/1pm5-hy02},
	note = {Version 1.0.0. RRID:SCR\_007345},
}

@article{hu_interpretable_2024,
title = {Interpretable medical image visual question answering via multi-modal relationship graph learning},
author = {Hu, Xinyue and Gu, Lin and Kobayashi, Kazuma and Liu, Liangchen and Zhang, Mengliang and Harada, Tatsuya and Summers, Ronald M. and Zhu, Yingying},
journal = {Medical Image Analysis},
volume = {97},
pages = {103279},
month = {October},
year = {2024},
doi = {10.1016/j.media.2024.103279},
url = {https://pubmed.ncbi.nlm.nih.gov/39079429/}
}

@article{goldberger_physiobank_2000,
	title = {{PhysioBank}, {PhysioToolkit}, and {PhysioNet}: {Components} of a new research resource for complex physiologic signals},
	author = {Goldberger, Ary L. and Amaral, Luis A. N. and Glass, Leon and Hausdorff, Jeffrey M. and Ivanov, Plamen C. and Mark, Roger G. and others},
	journal = {Circulation},
	volume = {101},
	number = {23},
	pages = {e215--e220},
	year = {2000},
	note = {Circulation [Online]. RRID:SCR\_007345},
}

@misc{medvlm_r1_2025,
    title = {MedVLM-R1: Incentivizing Medical Reasoning Capability of Vision-Language Models (VLMs) via Reinforcement Learning},
    author = {Pan, Jiazhen and others},
    year = {2025},
    month = {February},
    doi = {10.48550/arXiv.2502.19634},
    url = {https://arxiv.org/abs/2502.19634},
    note = {arXiv:2502.19634 [cs.CV]},
}

@misc{med_r1_2025,
    title = {Med-R1: Reinforcement Learning for Generalizable Medical Reasoning in Vision-Language Models},
    author = {Lai, Yuxiang and others},
    year = {2025},
    month = {March},
    doi = {10.48550/arXiv.2503.13939},
    url = {https://arxiv.org/abs/2503.13939},
    note = {arXiv:2503.13939 [cs.CV]},
}

@misc{rarl_2025,
	title = {RARL: Improving Medical VLM Reasoning and Generalization with Reinforcement Learning and LoRA under Data and Hardware Constraints},
	author = {Pham, Tan-Hanh and Ngo, Chris},
	year = {2025},
	note = {arXiv:2506.06600},
	url = {https://arxiv.org/abs/2506.06600}
	}

@misc{gmai_vl_r1_2025,
    title = {GMAI-VL-R1: Harnessing Reinforcement Learning for Multimodal Medical Reasoning},
    author = {Su, Yanzhou and Li, Tianbin and Liu, Jiyao and others},
    year = {2025},
    month = {April},
    doi = {10.48550/arXiv.2504.01886},
    url = {https://arxiv.org/abs/2504.01886},
    note = {arXiv:2504.01886 [cs.CV]},
}

@misc{med_rlvr_2025,
title = {Med-RLVR: Emerging Medical Reasoning from a 3B Base Model via Reinforcement Learning},
author = {Zhao, Zijian and others},
year = {2025},
note = {arXiv:2502.19655},
url = {https://arxiv.org/abs/2502.19655}
}

@misc{vision_r1_2025,
title = {Vision-R1: Evolving Human-Free Alignment in Large Vision-Language Models via Vision-Guided Reinforcement Learning},
author = {Zhan, Yufei and Zhu, Yousong and Zheng, Shurong and Zhao, Hongyin and Yang, Fan and Tang, Ming and Wang, Jinqiao},
year = {2025},
note = {arXiv:2503.18013},
url = {https://arxiv.org/abs/2503.18013}
}

@misc{reason_rft_2025,
title = {Reason-RFT: Reinforcement Fine-Tuning for Visual Reasoning},
author = {Tan, Huajie and Ji, Yuheng and Hao, Xiaoshuai and Lin, Minglan and Wang, Pengwei and Wang, Zhongyuan and Zhang, Shanghang},
year = {2025},
note = {arXiv:2503.20752},
url = {https://arxiv.org/abs/2503.20752}
}

@misc{gemex_thinkvg_2025,
title = {GEMeX-RMCoT: An Enhanced Med-VQA Dataset for Region-Aware Multimodal Chain-of-Thought Reasoning},
author = {Liu, Bo and Zhao, Xiangyu and He, Along and Chen, Yidi and Fu, Huazhu and Wu, Xiao-Ming},
year = {2025},
note = {arXiv:2506.17939},
url = {https://arxiv.org/abs/2506.17939}
}

@misc{effective_rl_medvqa_2025,
title = {Toward Effective Reinforcement Learning Fine-Tuning for Medical VQA in Vision-Language Models},
author = {Zhu, Wenhui and Dong, Xuanzhao and Li, Xin and Qiu, Peijie and Chen, Xiwen and Razi, Abolfazl and Sotiras, Aris and Su, Yi and Wang, Yalin},
year = {2025},
note = {arXiv:2505.13973},
url = {https://arxiv.org/abs/2505.13973}
}

@misc{mmedpo_2025,
title = {MMedPO: Aligning Medical Vision-Language Models with Clinical-Aware Multimodal Preference Optimization},
author = {Zhu, Kangyu and Xia, Peng and Li, Yun and Zhu, Hongtu and Wang, Sheng and Yao, Huaxiu},
year = {2025},
note = {arXiv:2412.06141, ICML 2025},
url = {https://arxiv.org/abs/2412.06141}
}

@misc{dpo_hallucination_2024,
title = {Direct Preference Optimization for Suppressing Hallucinated Prior Exams in Radiology Report Generation},
author = {Banerjee, Oishi and Zhou, Hong-Yu and Adithan, Subathra and Kwak, Stephen and Wu, Kay and Rajpurkar, Pranav},
year = {2024},
note = {arXiv:2406.06496},
url = {https://arxiv.org/abs/2406.06496}
}

@misc{litegpt_2024,
title = {LiteGPT: Large Vision-Language Model for Joint Chest X-ray Localization and Classification Task},
author = {Le-Duc, Khai and Zhang, Ryan and Nguyen, Ngoc Son and Pham, Tan-Hanh and Dao, Anh and Ngo, Ba Hung and Nguyen, Anh Totti and Hy, Truong-Son},
year = {2024},
note = {arXiv:2407.12064},
url = {https://arxiv.org/abs/2407.12064}
}

@misc{radvlm_2025,
    title = {RadVLM: A Multitask Conversational Vision-Language Model for Radiology},
    author = {Pellegrini, Chantal and Deperrois, Nicolas and others},
    year = {2025},
    month = {February},
    doi = {10.48550/arXiv.2502.03333},
    url = {https://arxiv.org/abs/2502.03333},
    note = {arXiv:2502.03333 [cs.CV]},
    publisher = {PhysioNet},
}

@misc{bahaaeldin2024nihchestxray14,
  author       = {BahaaEldin0},
  title        = {NIH-Chest-Xray-14},
  year         = {2024},
  publisher    = {Hugging Face},
  howpublished = {\url{https://huggingface.co/datasets/BahaaEldin0/NIH-Chest-Xray-14}},
  note         = {Accessed: 2025-12-25}
}

@inproceedings{Irvin2019CheXpert,
  title     = {CheXpert: A Large Chest Radiograph Dataset with Uncertainty Labels and Expert Comparison},
  author    = {Irvin, Jeremy and Rajpurkar, Pranav and Ko, Michael and Yu, Yifan and Ciurea-Ilcus, Silviana and Chute, Chris and Marklund, Henrik and Haghgoo, Behzad and Ball, Robyn and Shpanskaya, Katie and Seekins, Jayne and Mong, David A. and Halabi, Safwan S. and Lungren, Matthew P. and Ng, Andrew Y. and Langlotz, Curtis P.},
  booktitle = {Proceedings of the AAAI Conference on Artificial Intelligence},
  volume    = {33},
  number    = {01},
  pages     = {590--597},
  year      = {2019},
  doi       = {10.1609/aaai.v33i01.3301590},
  url       = {https://ojs.aaai.org/index.php/AAAI/article/view/3834}
}

@Misc{accelerate,
  title = {Accelerate: Training and inference at scale made simple, efficient and adaptable.},
  author = {Sylvain Gugger and Lysandre Debut and Thomas Wolf and Philipp Schmid and Zachary Mueller and Sourab Mangrulkar},
  howpublished = {\url{https://github.com/huggingface/accelerate}},
  year = {2022}
}

@misc{medgemma2024,
  title = {MedGemma: Medical Generative Multimodal Architecture},
  author = {{Google DeepMind}},
  year = {2024},
  howpublished = {\url{https://huggingface.co/google/medgemma-4b-it}},
  note = {Hugging Face model card}
}

@inproceedings{hu2022lora,
    title={Lo{RA}: Low-Rank Adaptation of Large Language Models},
    author={Edward J Hu and Yelong Shen and Phillip Wallis and Zeyuan Allen-Zhu and Yuanzhi Li and Shean Wang and Lu Wang and Weizhu Chen},
    booktitle={International Conference on Learning Representations},
    year={2022},
    url={https://openreview.net/forum?id=nZeVKeeFYf9}
}

@article{shao2024deepseekmath,
  title={DeepSeekMath: Pushing the Limits of Mathematical Reasoning in Open Language Models},
  author={Zhihong Shao and Peiyi Wang and Qihao Zhu and Runxin Xu and Junxiao Song and Xiao Bi and Haowei Zhang and Mingchuan Zhang and Y.K. Li and Y. Wu and Daya Guo},
  year={2024},
  url={https://arxiv.org/abs/2402.03300}
}

@article{liu2025understanding,
  title={Understanding {R1-Zero}-Like Training: A Critical Perspective},
  author={Zichen Liu and Changyu Chen and Wenjun Li and Penghui Qi and Tianyu Pang and Chao Du and Wee Sun Lee and Min Lin},
  journal={arXiv preprint arXiv:2503.20783},
  year={2025},
  url={https://arxiv.org/abs/2503.20783}
}

@misc{zheng2025stabilizing,
    title = {Stabilizing Reinforcement Learning with LLMs: Formulation and Practices}, 
    author = {Chujie Zheng and Kai Dang and Bowen Yu and Mingze Li and Huiqiang Jiang and Junrong Lin and Yuqiong Liu and An Yang and Jingren Zhou and Junyang Lin},
    year = {2025},
    eprint = {2512.01374},
    archivePrefix = {arXiv},
    primaryClass = {cs.LG},
    url = {https://arxiv.org/abs/2512.01374}
}

@misc{qwen2025qwen2.5vl3b,
    title = {Qwen2.5-VL-3B-Instruct},
    author = {Qwen Team},
    year = {2025},
    publisher = {Hugging Face},
    journal = {Hugging Face Model Hub},
    howpublished = {\url{https://huggingface.co/Qwen/Qwen2.5-VL-3B-Instruct}},
}

@inproceedings{compton2023,
  author    = {Compton, Rhys and Zhang, Lily and Puli, Aahlad and Ranganath, Rajesh},
  title     = {When More is Less: {I}ncorporating {A}dditional {D}atasets {C}an {H}urt {P}erformance {B}y {I}ntroducing {S}purious {C}orrelations},
  booktitle = {Proceedings of Machine Learning Research},
  volume    = {219},
  pages     = {1--24},
  year      = {2023},
  organization = {PMLR},
  eprint    = {2308.04431},
  archivePrefix = {arXiv},
  primaryClass = {cs.LG},
  url       = {https://arxiv.org/abs/2308.04431}
}

@article{murali2023,
  author    = {Murali, Nihal and Puli, Aahlad and Yu, Ke and Ranganath, Rajesh and Batmanghelich, Kayhan},
  title     = {Beyond {D}istribution {S}hift: {S}purious {F}eatures {T}hrough the {L}ens of {T}raining {D}ynamics},
  journal   = {Transactions on Machine Learning Research},
  year      = {2023},
  eprint    = {2302.09344},
  archivePrefix = {arXiv},
  primaryClass = {cs.LG},
  url       = {https://arxiv.org/abs/2302.09344}
}

@article{boland2025,
  author    = {Boland, Christopher and Tsaftaris, Sotirios A. and Dahdouh, Sonia},
  title     = {Preventing {S}hortcut {L}earning in {M}edical {I}mage {A}nalysis through {I}ntermediate {L}ayer {K}nowledge {D}istillation from {S}pecialist {T}eachers},
  journal   = {Machine Learning for Biomedical Imaging},
  volume    = {3},
  year      = {2025},
  eprint    = {2511.17421},
  archivePrefix = {arXiv},
  primaryClass = {cs.CV},
  doi       = {10.59275/j.melba.2025-8888},
  url       = {https://arxiv.org/abs/2511.17421}
}

@inproceedings{vogtlowell2023,
  author    = {Vogt-Lowell, Kevin and Lee, Noah and Tsiligkaridis, Theodoros and Vaillant, Marc},
  title     = {Robust {F}ine-{T}uning of {V}ision-{L}anguage {M}odels for {D}omain {G}eneralization},
  booktitle = {Proceedings of the IEEE High Performance Extreme Computing Conference},
  year      = {2023},
  eprint    = {2311.02236},
  archivePrefix = {arXiv},
  primaryClass = {cs.CV},
  url       = {https://arxiv.org/abs/2311.02236}
}

\end{document}